\documentclass[runningheads]{llncs}
\usepackage[T1]{fontenc}
\usepackage{graphicx}
\usepackage{subcaption}

\usepackage{cite}
\begin{document}
\title{Saliency-guided Emotion Modeling: Predicting Viewer Reactions from Video Stimuli}
%
%
\author{Akhila Yaragoppa\inst{1,2}\orcidID{0009-0001-4106-1868} \and
Siddharth\inst{1,3}\orcidID{0000-0002-1001-8218}}
\authorrunning{Yaragoppa and Siddharth}
\titlerunning{Saliency-guided Emotion Modeling}
%
\institute{HTI Lab, Plaksha University, Mohali, India \and
\email{akhila.yaragoppa@plaksha.edu.in} \and
\email{siddharth.s@plaksha.edu.in}\\
}
\maketitle              
\begin{abstract}
Understanding the emotional impact of videos is crucial for applications in content creation, advertising, and Human-Computer Interaction (HCI). Traditional affective computing methods rely on self-reported emotions, facial expression analysis, and biosensing data, yet they often overlook the role of visual saliency---the naturally attention-grabbing regions within a video. In this study, we utilize deep learning to introduce a novel saliency-based approach to emotion prediction by extracting two key features: saliency area and number of salient regions. Using the HD2S saliency model and OpenFace facial action unit analysis, we examine the relationship between video saliency and viewer emotions. Our findings reveal three key insights: (1) Videos with multiple salient regions tend to elicit high-valence, low-arousal emotions, (2) Videos with a single dominant salient region are more likely to induce low-valence, high-arousal responses, and (3) Self-reported emotions often misalign with facial expression-based emotion detection, suggesting limitations in subjective reporting. By leveraging saliency-driven insights, this work provides a computationally efficient and interpretable alternative for emotion modeling, with implications for content creation, personalized media experiences, and affective computing research.

\end{abstract}
\section{Introduction}
Understanding the emotional impact of videos and films on viewers has long been a subject of significant research interest due to its wide-ranging applications, including in advertising, video retrieval, and content summarization \cite{intro1}. Traditionally, researchers have relied on participant self-reported scores, facial expression analysis, electroencephalogram (EEG) recordings, and other physiological measurements to study the emotions elicited by watching videos. However, these approaches suffer from subjectivity, computational complexity, and a lack of focus on key visual elements. One critical gap in affective computing research is the role of visual saliency---the naturally attention-grabbing regions in a video that may significantly influence emotional responses.

This study utilized a deep learning model to detect salient regions in a video and introduces two novel interpretable features---``saliency area'' and ``number of salient regions''---derived from saliency maps of video content. Unlike traditional methods, these saliency-based features offer a new perspective on analyzing and predicting the emotional responses elicited by videos. To our knowledge, this is the first attempt to investigate the potential of using saliency features to correlate with viewer emotions and derive actionable insights for content creation. The biggest advantage of using saliency-based features over other methods that have previously been explored is that we would only need to look at the salient regions of the video to understand the emotion elicited. This would not only decrease the computational processing but will also aid content creators in figuring out what kind of salient features in the videos may induce which emotions in the users. To achieve this, we also utilize facial expression analysis and establish relationships between the saliency-based video features and users' reported emotions.

This research work makes the following key contributions:
\begin{enumerate}
    \item Video frames containing multiple salient regions are more likely to evoke emotions characterized by high valence and low arousal.
    \item Videos that evoke emotions with low valence and high arousal often focus on a single salient region at a time.
    \item Self-reports often misalign with facial expression-based emotions, suggesting limitations in subjective reporting.
\end{enumerate}

\section{Related Works}


\subsection{Video Saliency Prediction}
Saliency prediction is concerned with identifying the elements within a scene that naturally draw human attention. There are two main types of models for saliency prediction: saliency prediction models that estimate where observers will focus their gaze \cite{sal2}, and salient object detection models that identify objects of interest against a background \cite{sal3}. These models can further be classified into static saliency for images and dynamic saliency for videos.

Static saliency models have evolved from early hand-crafted features \cite{sal4} to more advanced CNN-based models that integrate deep learning techniques \cite{sal5,sal6,sal7} leading to better performance. The introduction of large datasets further improved their accuracy. Dynamic saliency models for videos address the additional challenge of capturing temporal changes. Early approaches adapted static models to video by analyzing frames independently, but these were soon outperformed by models designed to process spatial and temporal data simultaneously. Some of the leading models for dynamic saliency prediction are \cite{sal9,sal10,sal1}.

We use the $HD^2S$ model proposed in \cite{sal1}, a domain-agnostic architecture adaptable to diverse stimuli. Its key strength lies in generalizing across datasets without fine-tuning, enabled by gradient reversal layers that promote domain-independent feature learning. $HD^2S$ outperforms state-of-the-art methods on three of five metrics and ranks second-best on the remaining two in the DHF1K benchmark \cite{sal11}.

However, it may struggle with small objects or subtle motion and demands substantial computational resources at higher resolutions. The model is 116 MB in size with a runtime of 0.027 seconds. Further architectural and parameter details are available in \cite{sal1}.

\subsection{Facial Action Units Extraction}
Numerous methods exist for automatically identifying facial action units (AUs) from facial expressions based on the Facial Action Coding System (FACS) \cite{FACS_System}, using a range of machine learning and computer vision techniques to interpret facial behavior. Traditional approaches relied on handcrafted features to capture facial muscle movements \cite{facs6}, while recent deep learning methods leverage convolutional neural networks (CNNs) to learn features from large annotated datasets, improving the accuracy and robustness of AU detection \cite{facs7,facs5,facs4}.

We use the OpenFace library \cite{facs1}, a widely adopted open-source toolbox for facial behavior analysis, offering tools for facial landmark detection, head pose estimation, gaze tracking, and AU detection. Its AU detection module, based on \cite{facs2}, combines geometric and appearance-based features from video sequences to train machine learning models capable of real-time AU recognition.

Among the many available methods, we selected \cite{facs2} for its strong performance metrics and ease of integration.

\section{Methodology} 

\begin{figure}[h]
\centering
\includegraphics[width=0.65\textwidth]{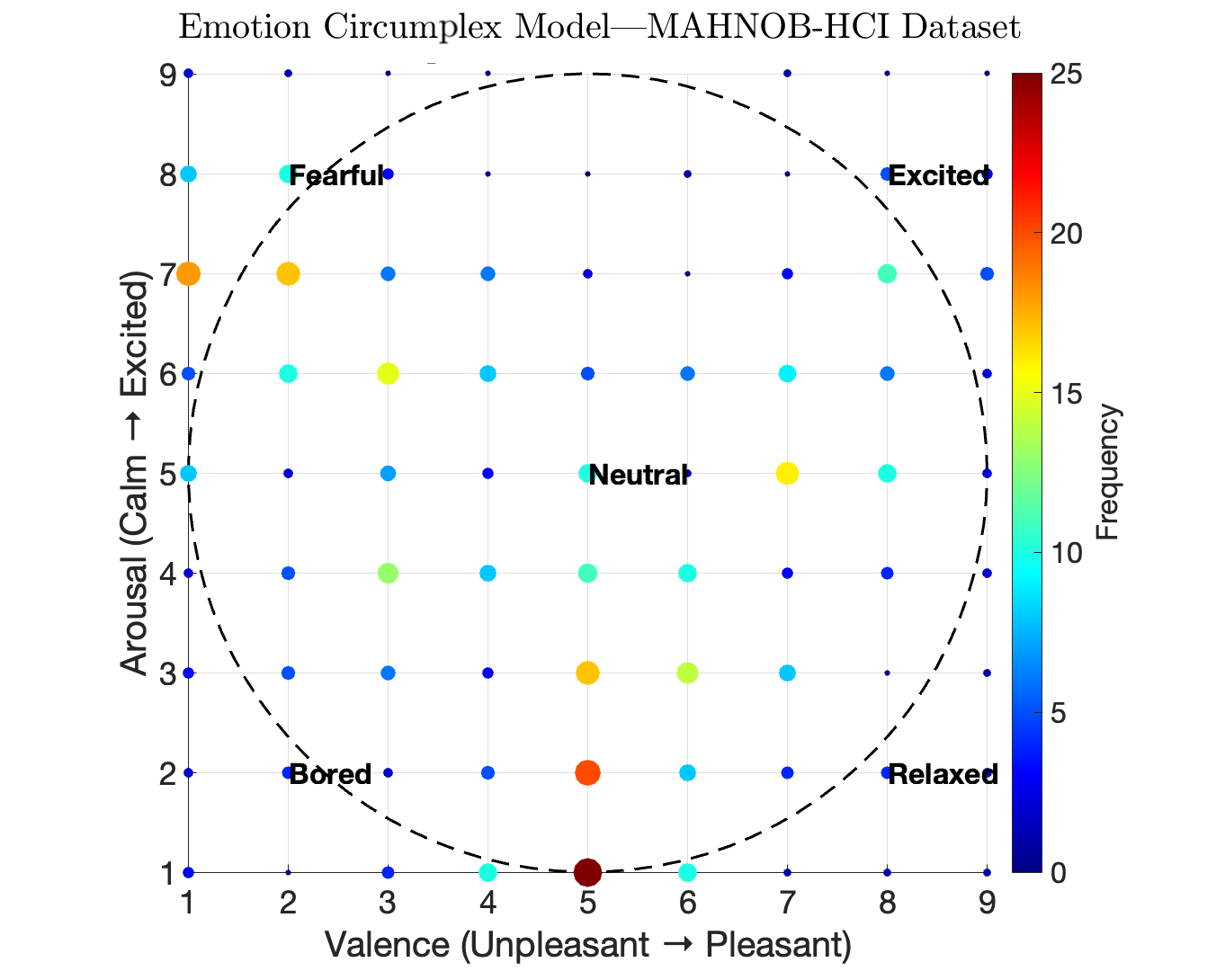}\hspace{2pc}%
\caption{Distribution of participants' self-reported emotions in the MAHNOB-HCI Dataset on the Emotion Circumplex Model}
\label{fig:va_model}
\end{figure}

\subsection{Dataset}
\textbf{Dataset and trials}: 
Several datasets have been gathered and publicly released to explore the link between elicitation videos and the emotions they provoke, providing data on participants' physiological responses, facial video recordings, and the specific videos they watched during the study. We choose the MAHNOB-HCI dataset \cite{mahnob} to conduct our analysis due to its wide adoption in the affective computing community. This dataset was collected from 30 participants watching 20 videos each. Excluding the trials with missing information, we finally use 527 trials, for which participants reported emotional responses (after each trial) and facial videos are available. We use the facial videos, felt valence score, felt arousal score, and the corresponding elicitation videos for our study. 

For each trial, we utilize the facial video, the corresponding valence and arousal labels reported by the participant, and the emotion elicitation video that the participant watched during the trial.

\textbf{Facial \& elicitation video processing}: We sample the frames in the video at a frequency of 2 frames per second, as the change in expressions in this duration is insignificant.

\textbf{Participant labels}: The valence and arousal labels reported by the participants are in the range 1-9 according to the emotion circumplex model \cite{russel}, with 1 representing low and 9 representing high valence/arousal (Figure \ref{fig:va_model}). 

\begin{figure}[h]
\centering
\includegraphics[width=0.8\textwidth]{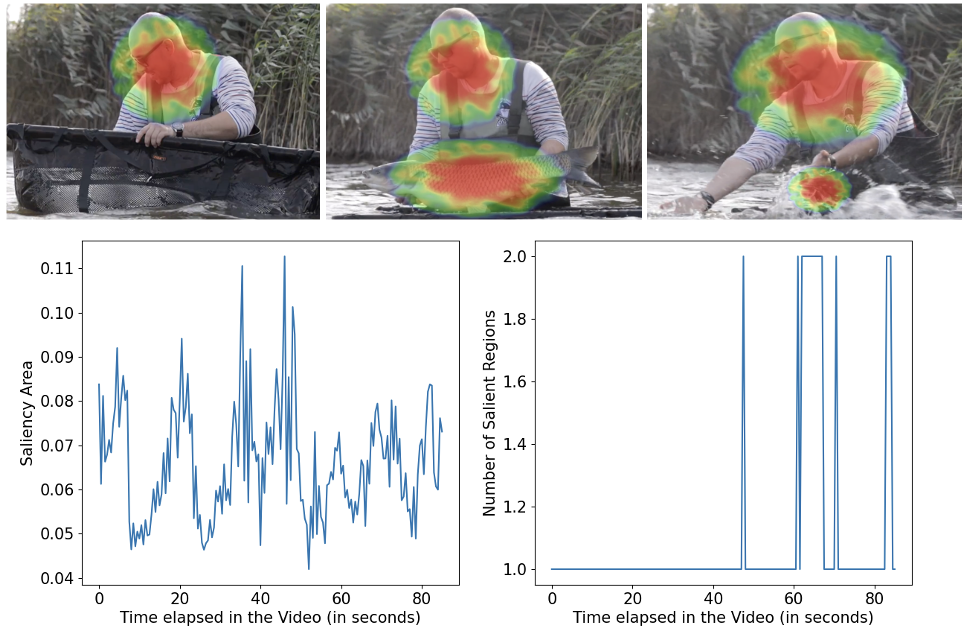}%
\caption{Features ``saliency area'' and ``number of salient regions'' (extracted using a deep neural network) vs. time for an example video stimulus (a few video frames and overlaid saliency heatmaps shown for reference) that the participants watch.}
\label{fig:sid_fig}
\end{figure}

\subsection{Saliency Features}
Extraction of saliency features is done in two steps:
\begin{itemize}
    \item [1.] A saliency video is passed through the $HD^2S$ model \cite{sal1}. The model outputs a saliency map for each frame in the video, with an intensity value for each pixel in the range of (0, 1). The deep learning model is run only for the preprocessed frames of the elicitation video, and the saliency map for each frame is generated and saved.
    \item [2.] Two saliency features (Figure \ref{fig:sid_fig}) are extracted from the saliency map output from the saliency model -- ``Saliency Area'' and ``Number of Salient Regions''. These features are extracted by computing the area and counts of regions generated by the saliency map outputs.
\end{itemize}

\textbf{Saliency Area} is calculated as the total area covered by all the regions that were identified as salient by the AI model. Since we have video inputs of varying sizes, we normalize this feature by dividing it by the total area of the video frame. The saliency area ranges from 0.02 to 0.13. This feature tells us what portion of the video frame is occupied by the salient objects(s). The saliency area predicted by the model is generally high when there are multiple salient regions in the video or when the salient objects cover a large area of the frame.

\textbf{Number of Salient Regions} is calculated as the number of distinct regions that were identified as salient by the AI model. This feature tells us the number of objects that the viewer is most likely to look at, in the video frame. We compute this using image processing on the saliency map to extract the number of contours in the map. In our dataset, number of salient regions is equal to one for most frames. It can increase up to 3 in some frames. The number of frames increases in the following scenarios:
1) when there is more than one region of interest in the video, and
2) when a scene change occurs the model takes a few frames to adjust to predict the accurate number of salient objects.

\subsection{Facial Action Units and Canonical Correlation Analysis}
We use the OpenFace \cite{facs1} library to extract Facial Action Units (AUs) from the preprocessed facial video frames. Specifically, we obtain the presence or absence of 18 AUs for each frame in the sequence. To examine the relationship between these AU features and video saliency features, we perform Canonical Correlation Analysis (CCA) \cite{cca}. This analysis identifies underlying correlations between the two sets of variables. The resulting CCA coefficient values are normalized such that their sum equals one, and we visualize these normalized values in each figure to show the relative contribution of each variable to the canonical components.


\section{Evaluation}

\subsection{Saliency Features and the Felt Emotions}
This section details the relationship between the extracted saliency features and the self-reported emotions felt by the participants. Each trial in the video has one valence and arousal score.

\begin{figure}[h]
\centering
\includegraphics[width=.8\linewidth]{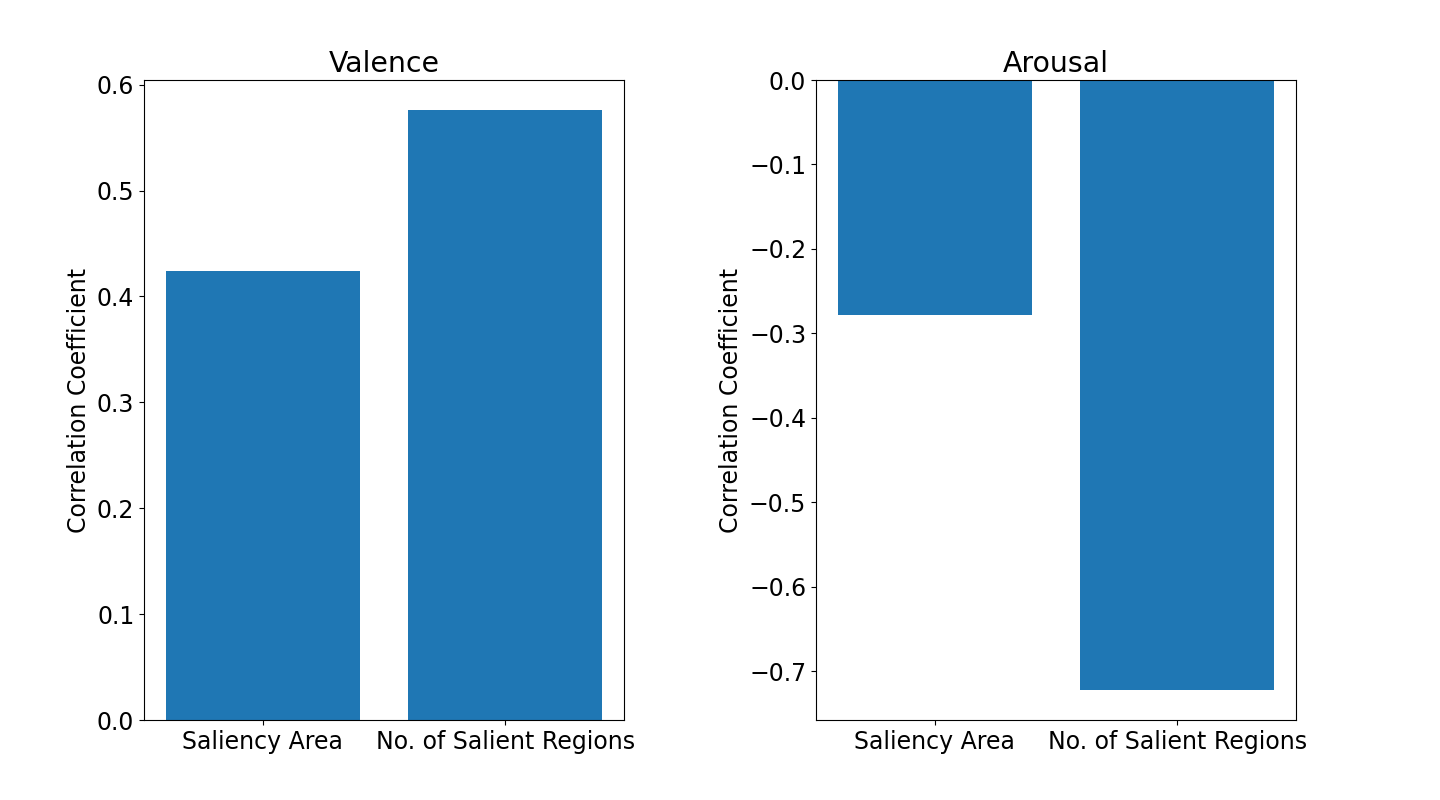}\hspace{2pc}%
\caption{Correlation between saliency features and the participant recorded emotions (Valence and Arousal).}
\label{fig:fig1}
\end{figure}

\begin{figure}
	\centering
	\begin{subfigure}{0.3\linewidth}
		\includegraphics[width=\linewidth]{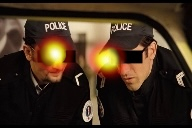}
		\caption{}
		\label{fig:subfigA}
	\end{subfigure}
	\begin{subfigure}{0.3\linewidth}
		\includegraphics[width=\linewidth]{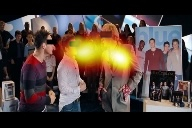}
		\caption{}
		\label{fig:subfigB}
	\end{subfigure}
	\begin{subfigure}{0.3\linewidth}
	        \includegraphics[width=\linewidth]{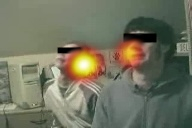}
	        \caption{}
	        \label{fig:subfigC}
         \end{subfigure}
	\caption{Example frames from a few visual stimuli having high valence and low arousal. The heatmap superimposed over the frame represents the salient regions identified by the deep learning network. As seen in (a) and (c) there can be multiple salient regions in a single frame.}
	\label{fig:subfigures}
\end{figure}

\begin{figure}
	\centering
	\begin{subfigure}{0.3\linewidth}
		\includegraphics[width=\linewidth]{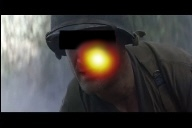}
		\caption{}
		\label{fig:subfigAAA}
	\end{subfigure}
	\begin{subfigure}{0.3\linewidth}
	        \includegraphics[width=\linewidth]{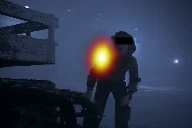}
	       \caption{}
	        \label{fig:subfigAAB}
         \end{subfigure}
         \begin{subfigure}{0.3\linewidth}
	        \includegraphics[width=\linewidth]{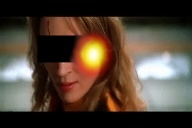}
	        \caption{}
	        \label{fig:subfigAAC}
         \end{subfigure}
	\caption{Example frames from a few visual stimuli having low valence and high arousal. The heatmap over the frame highlights salient regions identified by the deep learning network.}
	\label{fig:subfigures2}
\end{figure}

In Figure \ref{fig:fig1}, both saliency features show a positive correlation with self‐reported valence and a negative correlation with arousal, indicating that higher saliency areas and more salient regions align with high valence and low arousal. Figures \ref{fig:subfigures} and \ref{fig:subfigures2} illustrate these patterns in visual stimuli and corresponding saliency maps. Using Pearson’s Correlation Coefficient (PCC) \cite{pcc}, we found that the ``number of salient regions'' feature is highly negatively correlated with arousal (p = 0.0030), suggesting that high arousal typically corresponds to a focus on a single region. Other saliency features, however, exhibit only weak correlations with valence and arousal.

Figure \ref{fig:subfigures} shows representative frames from stimuli videos with high ``saliency area'' and multiple ``salient regions'', corresponding to a high mean valence score ($>5$) and low mean arousal score ($<5$). Most frames feature more than one salient region, resulting in a higher average saliency area. For these high-valence and low-arousal videos, it is observed that multiple shots in the video are of interactions between more than one character. There were also a few exceptions to this case, wherein there were multiple salient regions in the video, however, these videos had an average valence and low arousal reported.

Figure \ref{fig:subfigures2} shows representative frames from elicitation videos with a low ``saliency area'' and a single ``salient region'', corresponding to a low mean valence score ($<5$) and high mean arousal score ($>5$). Most frames focus on a single region, resulting in a lower average saliency area. For these low-valence and high-arousal videos it is observed that multiple frames in the video focus on a single character at a time, so you see only one salient character in the frame at a time. There were also a few exceptions to this case, wherein there was a single salient region detected, however the video had a high valence and low arousal reported.

Thus, Figures \ref{fig:fig1}, \ref{fig:subfigures}, and \ref{fig:subfigures2} offer key insights for content creators. To evoke high positivity and low arousal, multiple frames with several salient regions (e.g., interacting characters/objects) are effective. Conversely, visuals with fewer salient regions tend to elicit low positivity and high arousal, placing them in the low valence–high arousal quadrant of the emotion circumplex model.

\subsection{Saliency Features and Facial Action Units}
This section dives into the relationship between the extracted saliency features and the Facial Action Units (AUs) extracted from the participants' facial video. The emotions felt and self-reported by participants can be a culmination of many events other than simply the video being watched. For an accurate rating, the participants should have a neutral emotional baseline just before the experiment begins and should be focused on the visual stimuli while watching the video. The participant must also be capable of accurately judging and reporting how they ``feel'' after watching the video. To overcome these challenges, we also analyzed participants' facial expressions which may be a more reliable marker of the emotions they felt while watching the video rather than the self-reported scores at the end of the video trial. 

\begin{figure}[h]
\centering
\includegraphics[width=\linewidth]{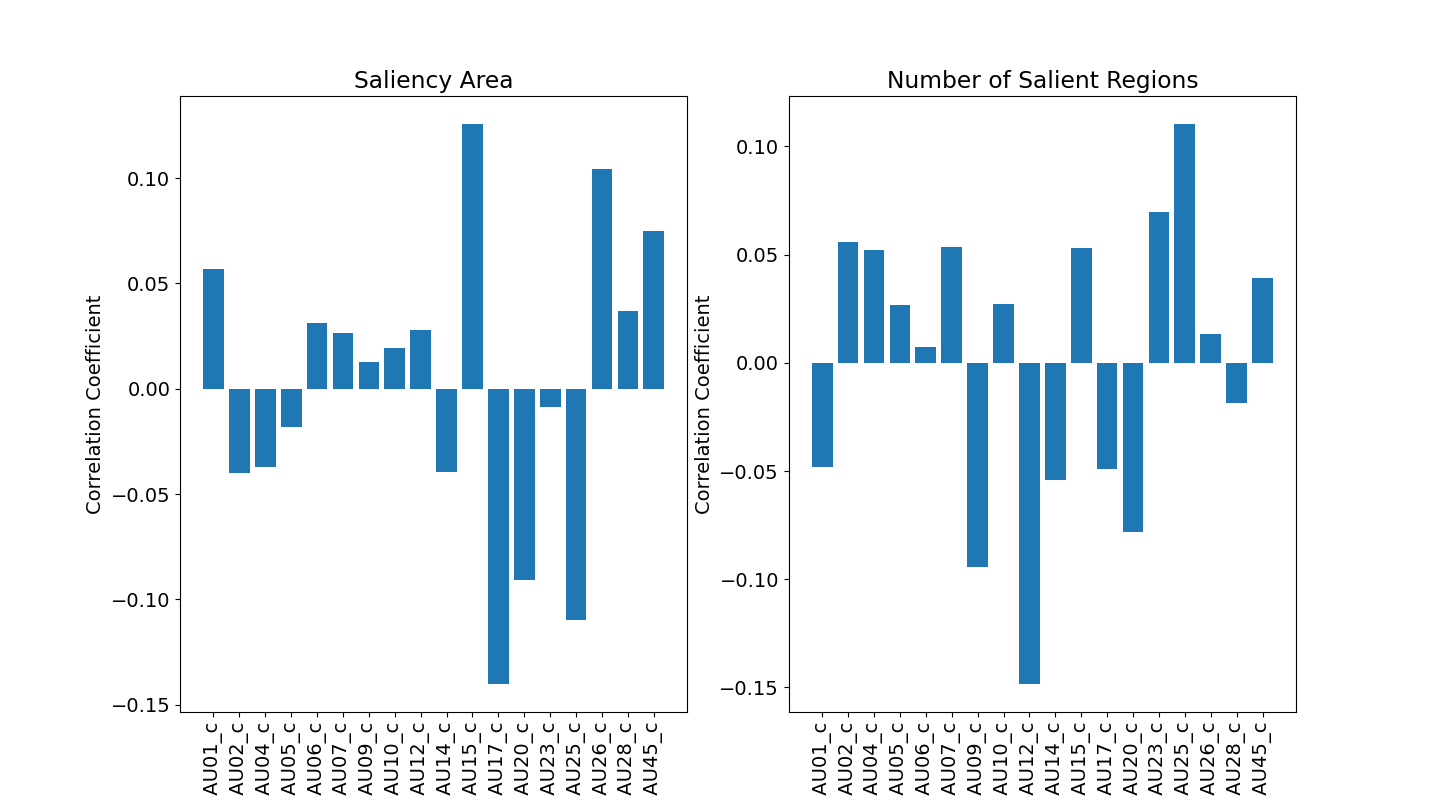}%
\caption{Correlation coefficients from CCA analysis of saliency features against detected AUs.}
\label{fig:fig2}

\end{figure}

\begin{figure}
	\centering
	\begin{subfigure}{0.45\linewidth}
            \centering
		\includegraphics[width=.85\linewidth]{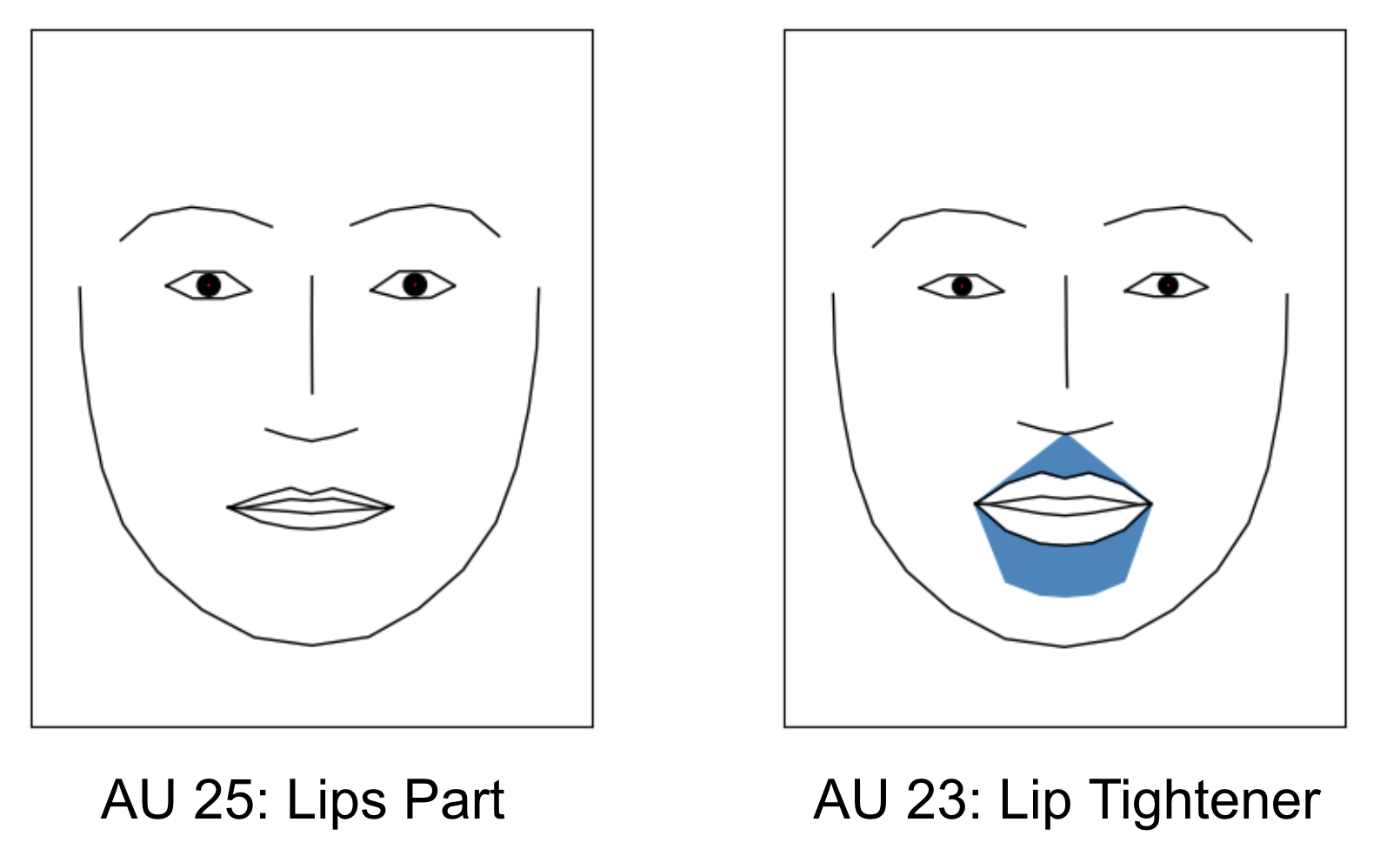}
		\caption{AUs contributing to more salient regions.}
		\label{fig:subfigA1}
	\end{subfigure}
         \begin{subfigure}{0.45\linewidth}
		\includegraphics[width=\linewidth]{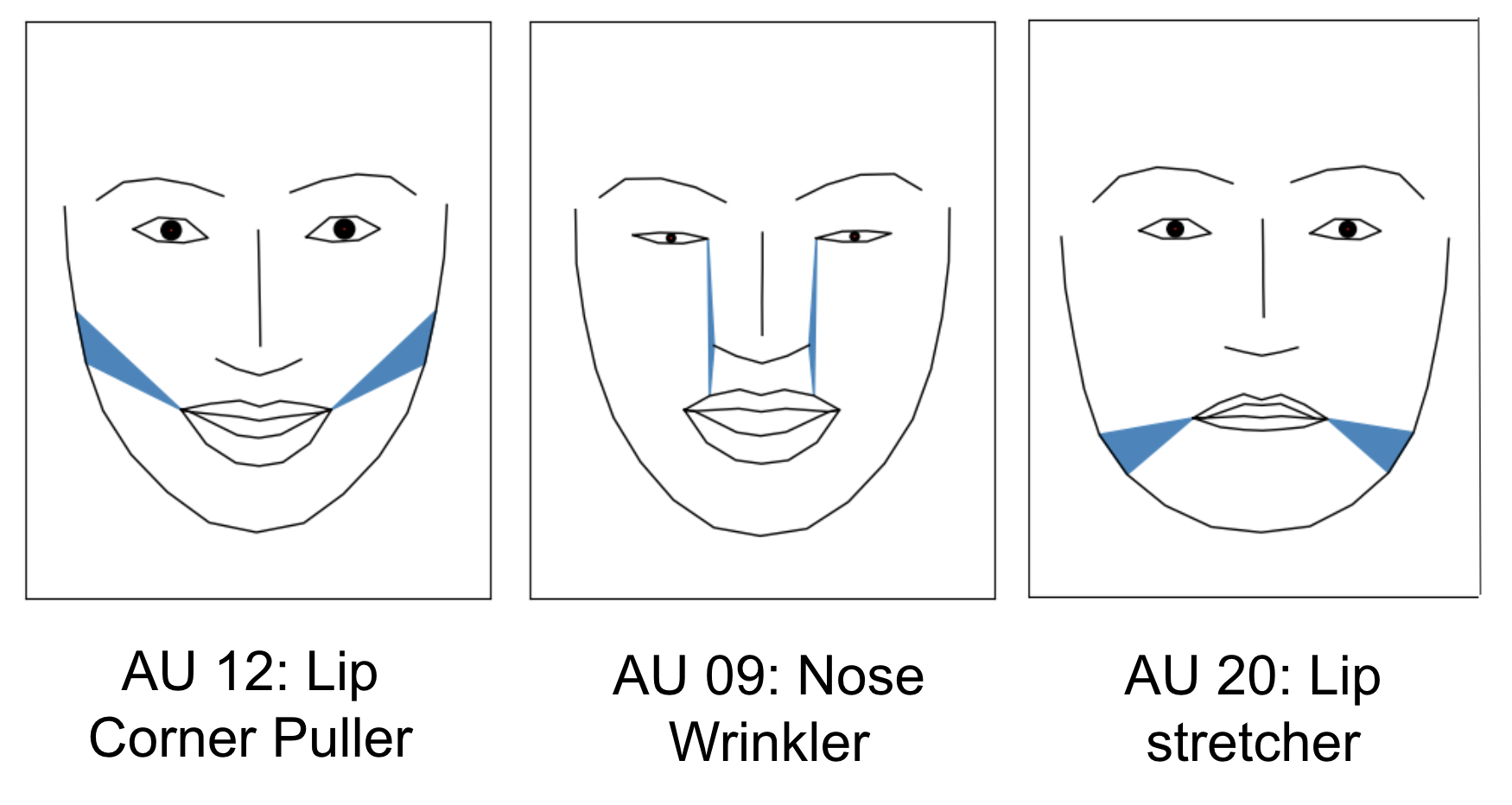}
		\caption{AUs contributing to fewer salient regions.}
		\label{fig:subfigB1}
	\end{subfigure}
    \caption{Top five Action Units contributing to Number of Salient Regions feature.}
    \label{fig:fig6}
\end{figure} 

\begin{figure}
	\centering
	\begin{subfigure}{0.45\linewidth}
            \centering
		\includegraphics[width=0.85\linewidth]{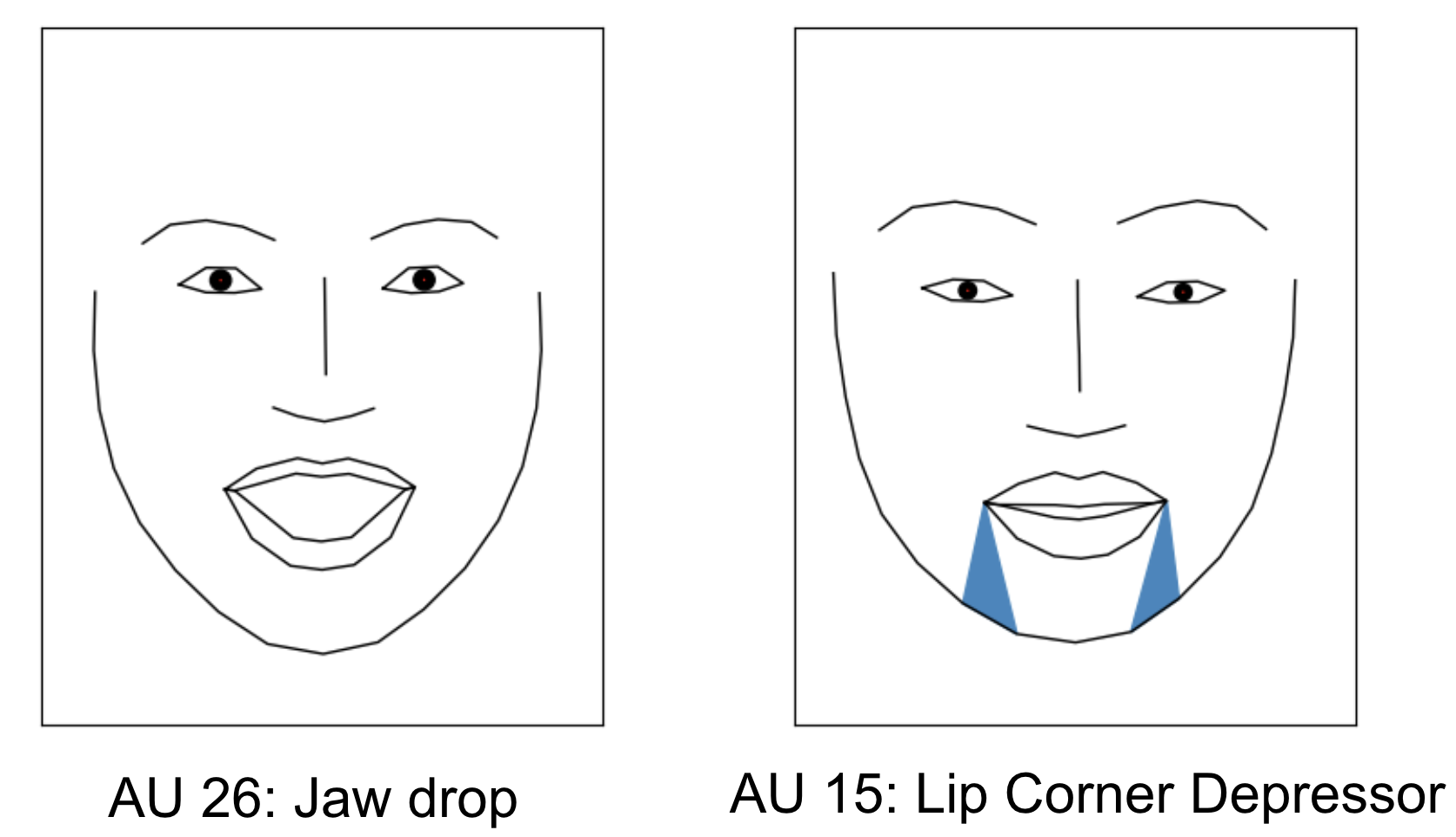}
		\caption{AUs contributing to high saliency area.}
		\label{fig:subfigAA}
	\end{subfigure}
         \begin{subfigure}{0.45\linewidth}
		\includegraphics[width=\linewidth]{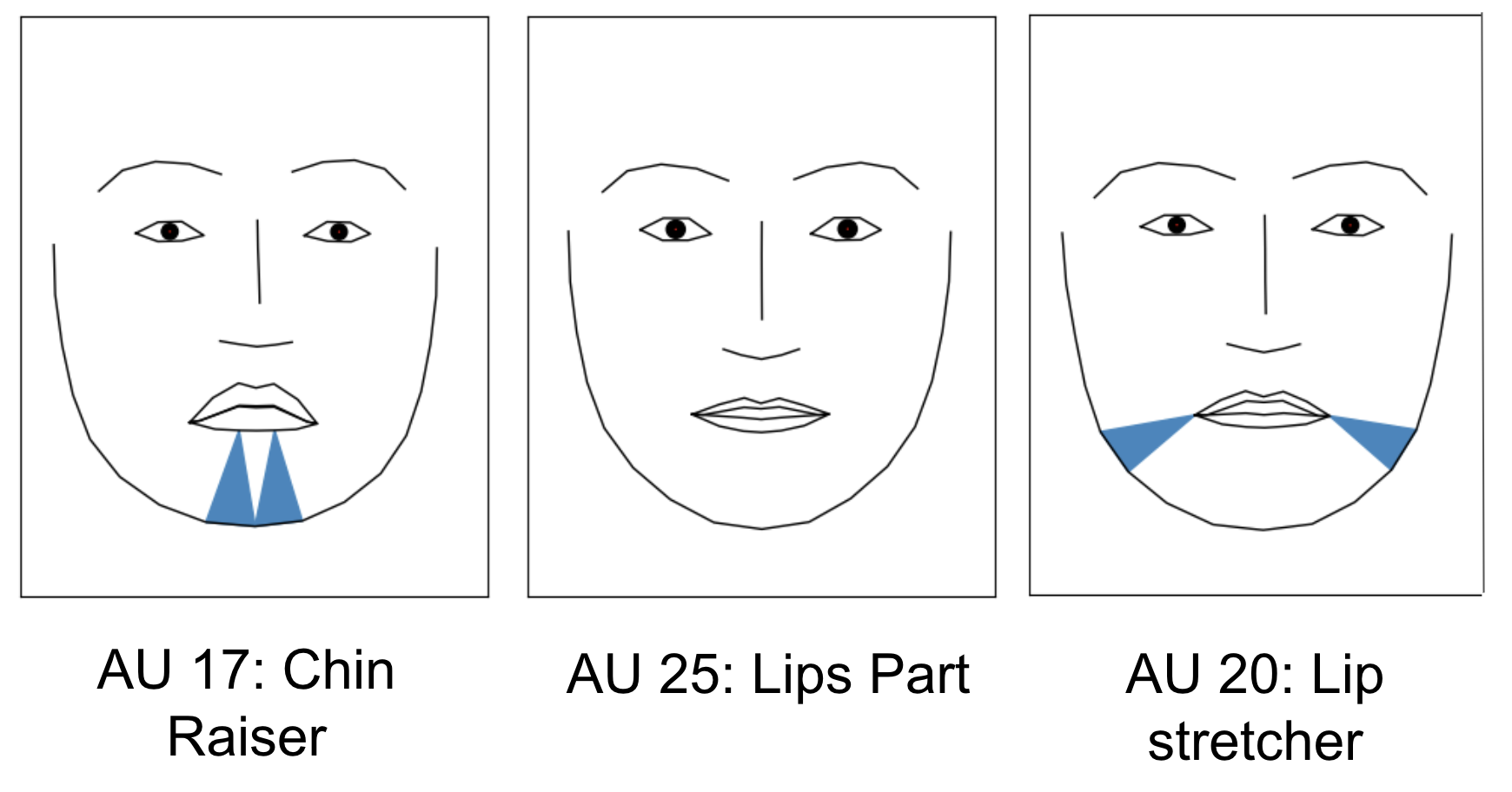}
		\caption{AUs contributing to low saliency area.}
		\label{fig:subfigAB}
	\end{subfigure}
    \caption{Top five Action Units contributing to Saliency Area feature.}
    \label{fig:fig7}
\end{figure}

Figures \ref{fig:fig2}, \ref{fig:fig6}, and \ref{fig:fig7} illustrate the relationship between facial AUs and video saliency features. Evidently, the top five AUs contributing to the ``saliency area'' feature (Figure \ref{fig:fig7}) are AU17 (Chin Raiser), AU15 (Lip Corner Depressor), AU25 (Lips Part), AU26 (Jaw Drop), and AU20 (Lip Stretcher), all concentrated in the mouth region (lips, chin, and jaws). Notably, AU15 and AU26 contribute positively to the ``saliency area'' feature. Prior literature associates AU15 with negative valence (sadness, fear, disgust) and AU26 with high arousal (surprise, fear). We infer that a high saliency area may indicate a higher likelihood of experiencing low valence and high arousal emotions.

Notably and quite interestingly, this is the opposite of what we observed above through the self-reported valence and arousal scores. This may mean that there could be a conflict between participants' self-reported emotions after watching the video stimulus and their facial expressions while watching it. This insight directly touches upon the ongoing debate about the best way to gauge user emotions in such emotion-invoking experiments and whether human physiological responses are more reliable indicators of human emotions than self-report \cite{debate}. Since facial expressions change temporally while participants watch a video stimulus, we propose that this debate could only be settled when participants are asked to continuously self-report their valence and arousal as the experiment progresses, not just at the end of each trial.

The top five AUs for the “number of salient regions” feature (Figure \ref{fig:fig6}) are AU12 (Lip Corner Puller), AU25 (Lips Part), AU09 (Nose Wrinkler), AU20 (Lip Stretcher), and AU23 (Lip Tightener), all concentrated in the lips and nose regions. Notably, AU25 and AU23 contribute positively---with AU23 linked to anger (a high-arousal emotion)---while AU12, AU09, and AU20 contribute negatively, being associated with contempt/joy, disgust, and fear, respectively. We infer that a low number of salient regions (i.e., a single region of interest) may indicate a higher likelihood of high-arousal emotions, consistent with the inferences made from the self-reported valence and arousal scores in the previous figure.

\subsection{Facial Action Units and the Felt Emotions}

\begin{figure}[h]
\centering
\includegraphics[width=\linewidth]{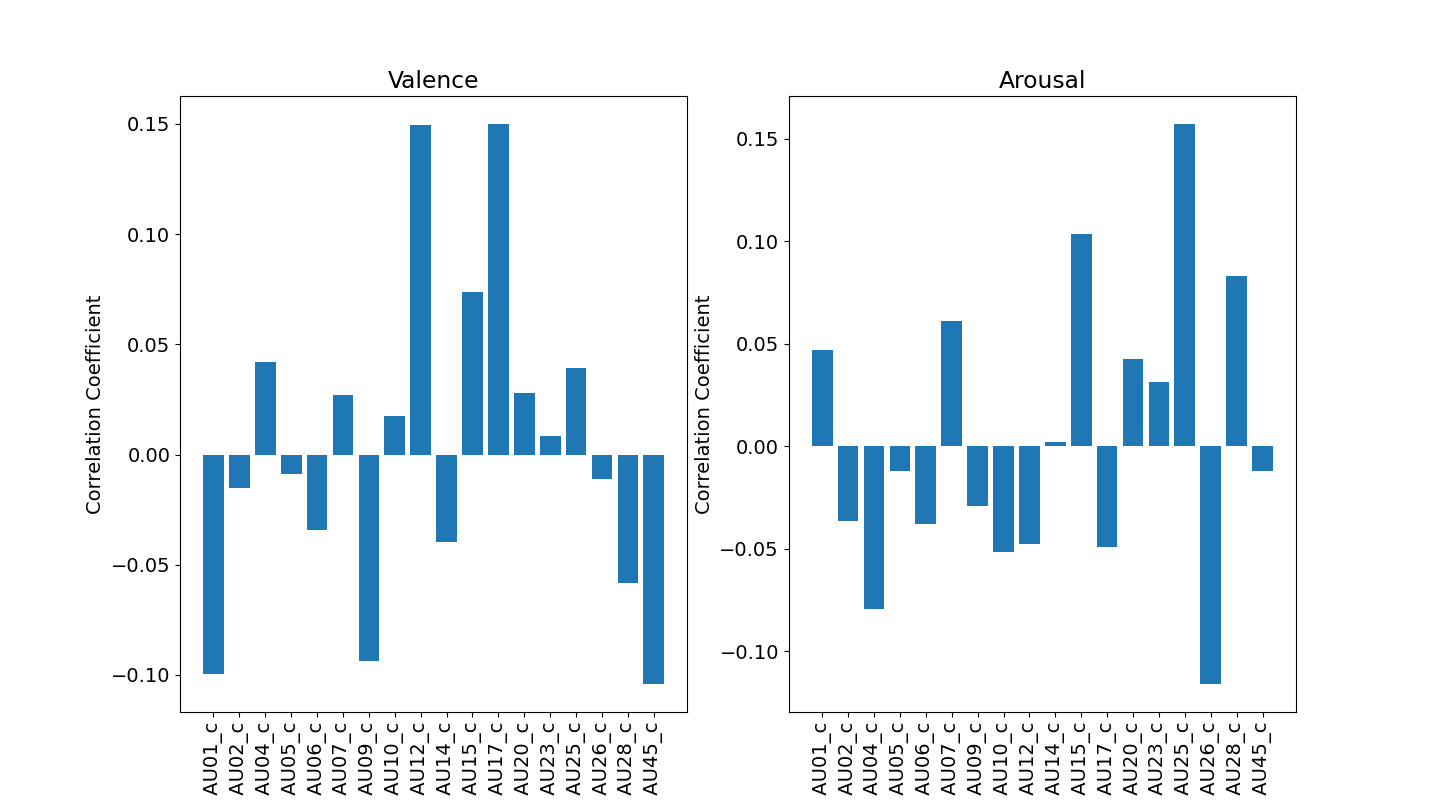}%
\caption{Correlation coefficients from CCA analysis of detected AUs against felt emotions.}
\label{fig:fig8}
\end{figure}

Finally, we plan to understand if analyzing facial expressions using AUs is consistent with self-reported valence and arousal. To achieve this, Figure \ref{fig:fig8} shows the coefficients from CCA between valence and arousal and a total of 18 AUs derived from participants' facial expressions. Thus, Figure \ref{fig:fig8} establishes the relationship between the emotions reported by the users and the emotions detected through the presence of AUs. We identify that a few of the AUs are highly correlated (positively or negatively) with user-reported valence and arousal. The top five AUs contributing to valence are: AU17 (Chin Raiser), AU12 (Lip Corner Puller), AU45 (Blink), AU01 (Inner Brow Raiser), and AU09 (Nose Wrinkler). Literature associates AU12—positively correlated with valence in our analysis—with joy, while AU01 and AU09—negatively correlated with valence—are linked to low-valence emotions such as sadness, surprise, fear, and disgust \cite{aumap}. 

The top five AUs for arousal are AU25 (Lips Part), AU26 (Jaw Drop), AU15 (Lip Corner Depressor), AU28 (Lip Suck), and AU04 (Brow Lowerer). AU25, AU15, and AU28 positively correlate with arousal, whereas AU26 and AU04 show negative correlations. Notably, AU15 has been linked to high-arousal emotions such as sadness and disgust, while AU26 and AU04 are associated with fear, surprise, and anger. Just like our analysis above, these results again show that, in general, facial expression analysis may not always be consistent with user-reported emotional valence and arousal. We believe that this is again true because of the fundamental issue with most such experimental protocols that ask participants to only report their emotions at the end of each trial while their facial expressions keep modulating through the trial.

\section{Conclusion}
This study examines how saliency-based features influence emotions elicited by video stimuli. We hypothesize that certain spatiotemporal regions have a stronger emotional impact and explore how facial expressions relate to self-reported emotions.

Our findings have broad implications. Researchers gain new insights into how video saliency shapes emotions, while content creators and marketers can better predict and guide audience responses. Discrepancies between facial expressions and self-reports suggest physiological responses may be more reliable emotion indicators.

However, a few limitations exist. Correlation does not imply causation, and external factors may influence emotions. Identifying ``emotionally charged” regions requires large, emotion-specific datasets for deep learning models. Future work will integrate these features to predict emotions across video segments rather than just correlations. Additionally, confounding factors, such as pre-experiment mood, personal connection, or recall accuracy, may have influenced self-reports. The assumption that facial expressions consistently reflect emotions may not always hold, especially when participants intentionally mask or modulate their expressions—introducing potential noise in the inference. The study also does not account for demographic variables such as age or gender, which could affect the generalisability of emotion predictions. Moreover, reliance on saliency features extracted from a single model (HD2S) may limit robustness, and future work could benefit from ensemble or comparative approaches. To further improve robustness and generalisability, future work could also involve conducting the same study on additional datasets to increase the diversity of findings.

Despite these challenges, our approach provides valuable insights into optimizing video content for more emotionally engaging media experiences.

\section*{Acknowledgments}
The authors are thankful to Harish and Bina Shah School AI \& CS and the Office of Research at Plaksha University for providing seed financial support through the Startup Research Grant Ref. No. OOR/PU-SRG/2023-24/06 for this research work.



%
%
%



\end{document}